\documentclass[12pt]{article}
\usepackage{wrapfig}
\usepackage{verbatim}

\usepackage{amsmath,amsfonts,bm}

\def\eqref#1{equation~\ref{#1}}

\def\1{\bm{1}}

\DeclareMathAlphabet{\mathsfit}{\encodingdefault}{\sfdefault}{m}{sl}
\SetMathAlphabet{\mathsfit}{bold}{\encodingdefault}{\sfdefault}{bx}{n}

\usepackage[left=2.5cm,
right=2.5cm,
top=2.3cm,
bottom=2.3cm,
headheight=20pt,
headsep=10pt,
footskip=25pt,
letterpaper]{geometry}
\usepackage[utf8]{inputenc}
\usepackage[T1]{fontenc}
\usepackage[english]{babel}
\usepackage{amsmath,amsfonts,amssymb,amsthm,thmtools}
\usepackage{graphicx}
\usepackage{hyperref}
\usepackage{fancyhdr}
\usepackage[normalem]{ulem}
\usepackage{graphicx}
\usepackage{stfloats}
\usepackage{wrapfig}
\usepackage{epstopdf}
\usepackage{cleveref}
\usepackage{subfloat}
\usepackage{subcaption}
\usepackage{xspace}
\usepackage{enumitem}
\usepackage{listings}
\usepackage{titlesec}
\usepackage{etoolbox}
\usepackage{setspace}
\usepackage{changepage}
\usepackage{etoolbox}
\usepackage{multirow}
\usepackage{booktabs}
\usepackage{tabularx}
\usepackage{wrapfig}
\usepackage{svg}
\usepackage[percent]{overpic}
\usepackage[round]{natbib}
\usepackage[colorinlistoftodos, shadow,color=blue!30!white
]{todonotes}
\usepackage{xpatch}
\usepackage{siunitx}

\renewcommand{\sfdefault}{phv}

\fancypagestyle{first}{\fancyfoot[R]{\small\thepage}}

\setlist[itemize]{leftmargin=1em,itemsep=0ex,topsep=0ex}
\titlespacing*{\paragraph}{0pt}{0ex plus .1ex}{1ex}
\titlespacing*{\section}{0ex}{2.3ex plus .3ex minus .0ex}{.6ex plus .3ex minus .2ex}
\titlespacing*{\subsection}{0ex}{1.5ex plus .3ex minus .5ex}{.4ex plus .2ex minus .1ex}
\titlespacing*{\subsubsection}{0ex}{1.2ex plus .3ex minus .3ex}{.3ex plus .2ex minus .2ex}

\xapptocmd\normalsize{%
\abovedisplayskip=.8em plus .2em minus .2em
\belowdisplayskip=.6em plus .1em minus .1em
\abovedisplayshortskip=.8em plus .2em minus .2em
\belowdisplayshortskip=.6em plus .1em minus .1em
}{}{}

\setcitestyle{numbers}
\renewcommand{\cite}[1]{\citep{#1}}

\definecolor{mydarkblue}{rgb}{0.0,0.15,0.7}
\hypersetup{%
colorlinks=true,
linkcolor=mydarkblue,
citecolor=mydarkblue,
filecolor=mydarkblue,
urlcolor=mydarkblue}

\makeatletter

  \renewcommand{\maketitle}{%
    \begingroup
      {\centering\LARGE\@title\par}%
      \vskip 1em
      \centering
      \begin{tabular}[t]{@{}c@{}}\strut\@author\strut\end{tabular}%
      \vskip 0.3in minus 0.1in
    \endgroup
  }
\makeatother

\title{HRM-Text: Efficient Pretraining Beyond Scaling}

\date{}
\author{
    \footnotesize
    Guan Wang$^{1,*,\dagger}$, %
    Changling Liu$^{1,*}$, %
    Chenyu Wang$^{2}$, %
    Cai Zhou$^{2}$, %
    Yuhao Sun$^{1}$, \\ %
    \footnotesize
    Yifei Wu$^{1}$, %
    Shuai Zhen$^{1}$, %
    Luca Scimeca$^{1}$, %
    Yasin Abbasi Yadkori$^{1,\dagger}$%
  \\[1ex]
  $^{1}$Sapient Intelligence
  \qquad
  $^{2}$MIT
}

\begin{document}
\pagestyle{fancy}

\maketitle
\thispagestyle{first}

\begingroup
\let\thefootnote\relax\footnotetext{$^{\dagger}$ Corresponding author. $^{*}$ Equal Contribution. Contact: \texttt{research@sapient.inc}. 

Code available at: \href{https://github.com/sapientinc/HRM-Text}{\texttt{github.com/sapientinc/HRM-Text}}}
\endgroup

\vspace{-2.3em}
\begin{abstract}
\vspace{-0.7em}
The current pretraining paradigm for large language models relies on massive compute and internet-scale raw text, creating a significant barrier to foundational research. In contrast, biological systems demonstrate highly sample-efficient learning through multi-timescale processing, such as the functional organization of the frontoparietal loop. Taking this as inspiration, we introduce HRM-Text, which replaces standard Transformers with a Hierarchical Recurrent Model (HRM) that decouples computation into slow-evolving strategic and fast-evolving execution layers. To stabilize this deep recurrence for language modeling, we introduce \textit{MagicNorm} and warmup deep credit assignment. Furthermore, instead of standard raw-text pretraining, we train exclusively on instruction-response pairs using a task-completion objective and PrefixLM masking. Serving as an empirical existence proof of efficient pretraining, a 1B-parameter HRM-Text model trained from scratch on only 40 billion unique tokens and \$1,500 budget achieves 60.7\% on MMLU, 81.9\% on ARC-C, 82.2\% on DROP, 84.5\% on GSM8K, and 56.2\% on MATH. Despite utilizing roughly 100-900x fewer training tokens and 96-432x less estimated compute than standard baselines, HRM-Text performs competitively with 2--7B parameter open models. These results demonstrate that co-designing architectures and objectives can radically reduce the compute-to-performance ratio, making pretraining from scratch accessible to the broader research community.

\vspace{-1em
}
\end{abstract}

\begin{figure}[h]
  \centering
  \includegraphics[width=0.99\linewidth]{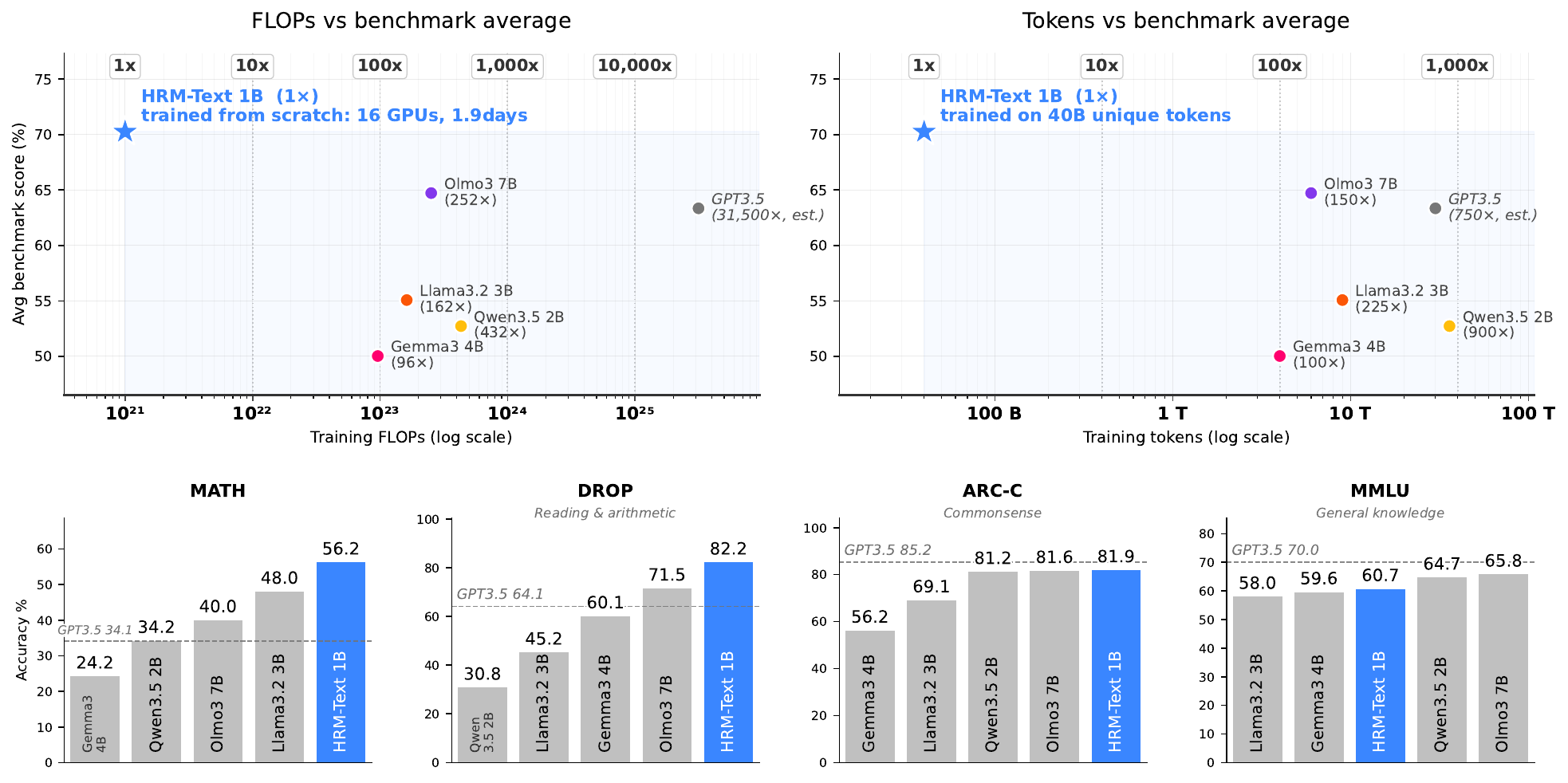}
  \vspace{-0.4em}
    \caption{\textbf{Pretraining efficiency.} Trained from scratch in 1.9 days on 16 GPUs, HRM-Text 1B achieved performance competitive with substantially larger 2--7B foundation models while utilizing up to 432$\times$ less compute and 900$\times$ fewer training tokens.}
  \label{fig:benchmark_bars}
  \vspace{-1em}
\end{figure}

\clearpage

\section{Introduction}

The remarkable success of large language models (LLMs) is currently driven by a monolithic recipe: massive, multi-stage pipelines that begin with broad unsupervised pretraining over internet-scale raw text. While undeniably effective, this brute-force scaling paradigm is highly inefficient in data-limited regimes. Massive compute is spent predicting prompt-like or task-irrelevant text simply to build generalized  representations~\citep{kaplan2020scaling,hoffmann2022training,tay2023ul2}. Consequently, this extreme computational barrier has largely locked the broader research community out of foundational pretraining exploration. The prevailing assumption is that without immense compute clusters and trillions of tokens, investigating new architectures or training from scratch is futile.

This brute-force data hunger stands in stark contrast to human intelligence, which can grasp governing rules and perform heuristic-guided search from only a few examples. In our previous work, we introduced the Hierarchical Recurrent Model (HRM), a dual-timescale architecture inspired by the functional organization of the biological frontoparietal loop~\citep{HRM2025}. By decoupling deliberation into a slow-evolving strategic layer and a fast-evolving execution layer, HRM provided a structural inductive bias that helped avoid local stagnation and successfully guided symbolic search on combinatorial tasks.

However, scaling recurrent architectures to the open-ended complexities of language modeling introduces severe gradient-instability risks~\citep{bengio1994learning,chowdhury2025investigating,hutchins2022blockrecurrent,zucchet2024recurrent}. A structural prior alone is insufficient; achieving competitive open-domain performance requires a holistic codesign. In this paper, we demonstrate that architecture and training methods are profoundly important once again. We explore two major, synergistic directions to realize this sample-efficient engine:

\begin{itemize}
    \item \textbf{Architectural Exploration:} To achieve deep computation without a proportional explosion in parameter counts, we build upon HRM's modular, multi-timescale recurrence. The fast $L$-module performs local iterative refinement, while the slow $H$-module maintains stable semantic context across cycles~\citep{HRM2025}. To make this deep recurrence mathematically viable for language, we introduce stabilization techniques like \textit{MagicNorm} and warmup deep credit assignment, which bound forward activation variance while maintaining backward optimization stability~\citep{xiong2020layer,liu2020understanding,tallec2017unbiasing}.
    
    \item \textbf{Objective Exploration:} We challenge the dogma of autoregressive pretraining on raw text. Since models are primarily used for conditional generation at inference time, we pretrain HRM-Text directly from scratch on instruction-response pairs~\citep{wei2022finetuned,sanh2022multitask,longpre2023flan}. We optimize a task-completion objective, computing the negative log-likelihood loss exclusively over the response: $-\log P(x_a \mid x_q)$~\citep{sutskever2014,raffel2020exploring,sanh2022multitask}. We pair this with a PrefixLM attention mask, which allows full bidirectional (encoder-like) attention across the instruction tokens while preserving standard causal generation for the response~\citep{liu2018generating,dong2019unified,raffel2020exploring,tay2023ul2}.
\end{itemize}

When these two directions are combined, the result is an empirical existence proof that defies the current scaling dogma. Trained from scratch on a low budget of only 40B unique tokens, HRM-Text achieves strong performance on most benchmarks against contemporary open models like Llama, Qwen,  Gemma, OLMo, Ouro and Huginn~\citep{meta2024llama3,yang2025qwen3,gemmateam2025gemma3technicalreport,olmo2025olmo,zhu2025scaling,geiping2026scaling}. Strikingly, it reaches this performance neighborhood using roughly $100\text{-}900\times$ fewer training tokens and $96\text{-}432\times$ less estimated training compute than these baselines, as shown in \Cref{fig:benchmark_bars} and \Cref{tab:eval_open_weight}.

We do not present HRM-Text as the final or optimal language model, but rather as proof that specific structural priors and targeted training objectives can radically alter the compute-to-performance ratio. Because the entry price is vastly reduced, this methodology democratizes foundational AI research. Pretraining from scratch is accessible again—we invite the community to join us in exploring how far smart architectures and focused objectives can go.

\section{Methods}

\label{sec:hrm_arch}

\begin{figure}[h]
  \centering
  \includegraphics[width=\linewidth]{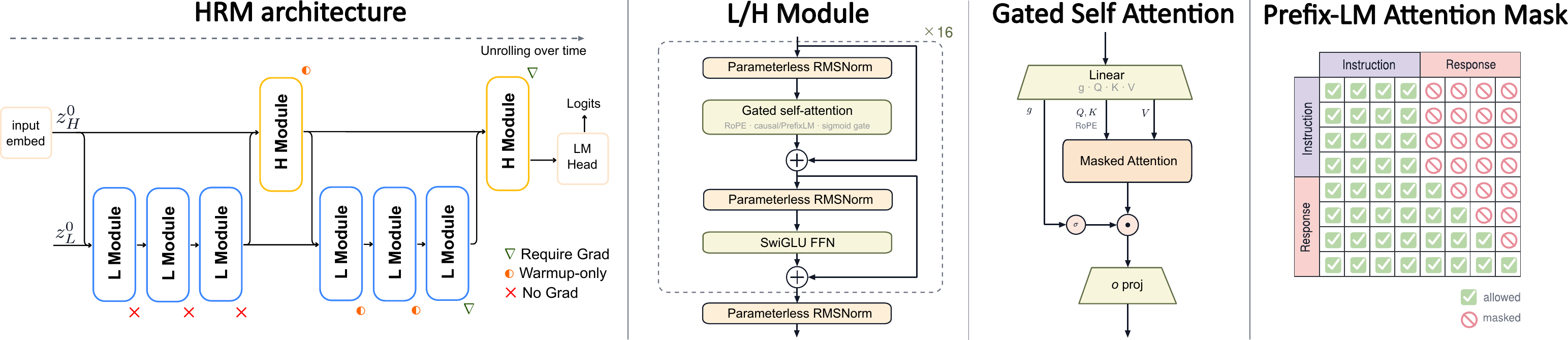}
  \caption{\textbf{HRM-Text architecture.} (a) Dual-timescale recurrent design comprising L and H modules. (b) L/H module internals featuring \textit{MagicNorm}---PreNorm blocks followed by final norm. (c) Sigmoid-gated multi head self-attention. (d) PrefixLM mask enabling bidirectional attention on instruction.}
  \label{fig:hrm_architecture}
\end{figure}

HRM-Text builds upon an improved HRM architecture, featuring a dual-timescale recurrence~\cite{HRM2025}. The forward pass is initialized with a high-level state, $z_H^0$, derived from the input token embeddings, alongside a fixed low-level state, $z_L^0$. The core processing sequence consists of two high-level cycles. Each cycle executes three fast $L$ module updates followed by a single slow $H$ module update. Token logits are generated by applying a linear head to the output of the final $H$ module state. We employ a warmup deep credit assignment strategy: gradients are initially backpropagated through only the final two recurrent steps, expanding to the final five steps as training progresses.

Internally, both the $H$ and $L$ recurrent modules are structured using \textit{MagicNorm}. Additionally, we utilize parameterless RMSNorm (omitting the learnable $\gamma$ parameter)~\cite{zhang2019root}, SwiGLU activation functions~\cite{shazeer2020glu}, Rotary Position Embeddings (RoPE)~\cite{su2024roformer}, and a sigmoid-gated self-attention mechanism~\cite{qiu2026gated}.

In contrast to standard autoregressive pretraining on raw text, we optimize a task-completion objective. The model is pretrained directly on instruction-response pairs $(x_q, x_a)$ from scratch using a negative log-likelihood (NLL) loss computed exclusively over the response, $-\log P(x_a|x_q)$. This objective is naturally paired with a PrefixLM attention mask, enabling full bidirectional attention across the instruction tokens.

In the following sections, we detail the specific mechanics that enable HRM-Text's extreme efficiency. Section \ref{subsec:stable} delves into  our novel stabilization techniques, while Section \ref{sec:task_obj} explores the task-completion pretraining objective and PrefixLM masking strategy.

\subsection{Scaling to language with recurrence}\label{subsec:stable}

\subsubsection{Stabilization via \textit{MagicNorm}}

Although the original HRM demonstrated strong performance on symbolic tasks, scaling recurrent architectures to language modeling introduces severe gradient-instability risks. Transformer design already involves a compromise in the placement of normalization layers\citep{xiong2020layer, liu2020understanding}; recurrence amplifies this compromise because the same transformation is repeatedly applied over many steps.

\textbf{PostNorm} \citep{vaswani2017attention} places the normalization outside the residual branch:
$$ h_{l} = \text{Norm}(h_{l-1} + \text{Sublayer}(h_{l-1})) $$
This effectively bounds activation variance and can improve expressivity, but it disrupts the clean identity path and can lead to vanishing gradients in deeper networks \citep{liu2020understanding}.

\textbf{PreNorm} places the normalization inside the residual branch:
$$ h_{l} = h_{l-1} + \text{Sublayer}(\text{Norm}(h_{l-1})) $$
This maintains a direct identity path, $h_L = h_0 + \sum_{l=1}^L \text{Sublayer}(\cdot)$, allowing gradients to flow more directly to early layers. However, the unnormalized residual accumulation can cause hidden-state variance to grow with depth, which may lead to representation collapse or reduced performance relative to PostNorm.

\textbf{\textit{MagicNorm}:} To address this tradeoff in recurrent models, we introduce \textit{MagicNorm}, which exploits the asymmetry between the forward and backward computational horizons induced by truncated backpropagation through time (TBPTT).

Let $N$ denote the total number of recurrent forward steps and $K$ denote the truncated backward horizon, where $K \ll N$. In \textit{MagicNorm}, each recurrent module is composed of $L$ internal PreNorm blocks, but is capped with a final  normalization layer at its exit:
$$ z_{n} = \text{Norm}\left(z_{n-1} + \sum_{l=1}^L \text{Sublayer}_l(\text{Norm}(\cdot))\right) $$

During the \emph{forward pass}, the recurrent state $z$ is subjected to $N$ module-level normalization operations. Because these norms sit directly on the main recurrent pathway, they bound activation variance at the end of every recurrent step. This prevents the unbounded variance growth of pure PreNorm and gives the recurrent core PostNorm-like forward stability.

Conversely, during the \emph{backward pass}, the truncated gradient horizon means the error signal passes through the module-level normalization only $K$ times. Within that same horizon, the gradient also flows through $L$ internal PreNorm identity connections. Since $K$ is small relative to the full recurrence depth $N$, MagicNorm behaves more like a stable PreNorm architecture during optimization.

\subsubsection{Warmup deep credit assignment}

The original HRM uses a fixed 1-step gradient strategy, backpropagating only through the last two recurrent steps (last $H$ and last $L$). We extend this approach with \textit{warmup deep credit assignment}. The schedule is motivated by temporal-curriculum principles: early optimization is restricted to short credit-assignment paths, and longer paths are introduced only after the model has reached a more stable regime. This design is also consistent with biological accounts of temporal learning, where local traces can support delayed credit assignment~\citep{izhikevich2007distal}, reward-predictive signals can shift from reward-proximal events to earlier cues~\citep{amo2022gradual}, and developmental curricula can improve sequence learning by exposing learners to shorter-range structure before longer-range dependencies~\citep{elman1993starting}.

Operationally, we dynamically adjust the backward gradient horizon, $K$. During early pretraining, we compute gradients through only the last two recurrent steps ($K=2$), then linearly warm up the horizon to the last five steps ($K=5$). This progressive deepening allows the model to exploit longer recurrent computation while reducing exposure to the optimization pathologies that often arise from long gradient paths at initialization. Because the warmup phase backpropagates through fewer recurrent steps than the final setting, it also reduces the average backward-pass computation and accelerates early training.

\subsection{Task-completion objective and PrefixLM}
\label{sec:task_obj}

The dominant paradigm for training foundation models relies on a resource-intensive, multi-stage pipeline. From T5 through modern large language models~\cite{raffel2020exploring}, training typically begins with broad unsupervised pretraining and is followed by higher-quality mid-training.

In the pretraining phase, models are trained on internet-scale raw corpora to learn general language representations. In the mid-training (or annealing) phase, the model is refined on high-quality text, usually instruction-like data. In both phases, the model optimizes an NLL objective over all tokens
$$-\log P(x)$$

While effective, this approach can be inefficient in the data- and resource-limited regime. Broad raw-text pretraining consumes most of the compute and data, and much of the token-level loss is spent on predicting prompt-like or task-irrelevant text. Yet at inference time, models are applied primarily on conditional generation: \textit{given a query or instruction, they must produce an appropriate response.}

To improve sample efficiency, HRM-Text omits broad raw-text pretraining and trains exclusively on instruction-response pairs from scratch. Given an example containing an instruction and response $x = (x_q, x_a)$, we optimize the NLL of the response conditioned on the instruction:
$$-\log P(x_a|x_q)$$

\begin{figure}[!t]
  \centering
  \includegraphics[width=\linewidth]{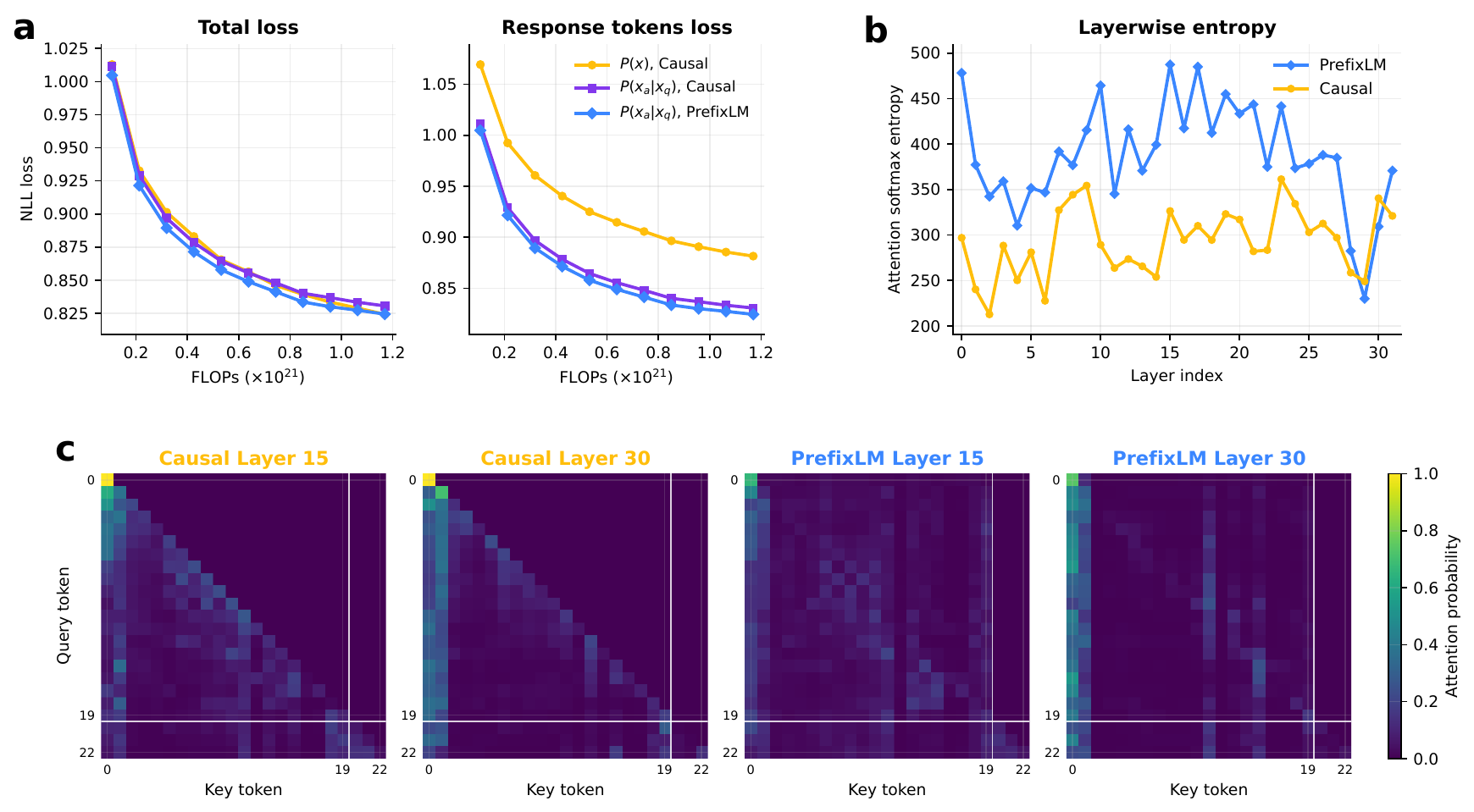}
  \caption{\textbf{Task-completion and PrefixLM improve response modeling.} (a) Compared with full causal language modeling $P(x)$, response-only training $P(x_a|x_q)$ lowers response-token NLL. PrefixLM further improves response loss. (b) PrefixLM increases layerwise attention entropy relative to causal masking, suggesting broader use of the prompt. (c) Attention maps illustrate the qualitative difference: causal attention remains mostly local and triangular, while PrefixLM enables global bidirectional interactions among prompt.}
  \label{fig:obj_loss_and_attention}
\end{figure}

By not predicting the instruction tokens, the model concentrates its parameter updates on generating accurate responses. \Cref{fig:obj_loss_and_attention}-(a) illustrates this effect. Although the total loss is comparable with and without the task-completion objective, the error associated with the response component is substantially lower.

Furthermore, this single-stage conditional objective naturally aligns with a PrefixLM attention mask \citep{raffel2020exploring}. Because the model is never required to autoregressively predict the instruction $x_q$, we remove the causal masking over the instruction segment: all instruction tokens attend to one another bidirectionally, while standard causal masking is maintained over the response sequence. This gives HRM-Text an encoder--decoder-like separation inside a decoder-style implementation. The instruction segment is first integrated as a fully visible context, analogous to an encoder-side representation, while the response segment is generated autoregressively, analogous to a decoder.

\Cref{fig:obj_loss_and_attention} (b) shows that PrefixLM leads to higher attention softmax entropy, indicating attention over a more diverse set of tokens. \Cref{fig:obj_loss_and_attention} (c) shows that causal attention is more localized, whereas PrefixLM attention is more global and diverse. Together, the response-only conditional loss and PrefixLM attention improve sample efficiency in the data- and compute-restricted regime.

\section{Results} \label{sec:results}

As the central question of this paper is whether a model trained from random initialization under a small pretraining budget can reach a meaningful open-model performance regime, we approach this question as a small-budget design exploration: first, whether architectural choices can improve the use of fixed training compute, and second, whether the objective and input structure can increase the yield of each training example. Finally, we compare HRM-Text with contemporary fully open and open-weight models to quantify its efficiency relative to current pretraining practice, and analyze whether the recurrent architecture increases effective depth. Training details for all models are provided in \Cref{sec:training_details}.

Across these experiments, HRM-Text is trained from scratch on the task-formatted mixture described in \Cref{sec:methods:data}, using only 40B unique tokens. We report all the performance from a single HRM-Text checkpoint.

\subsection{Architecture efficiency under matched training compute}

The first part of this exploration asks how much architecture design can improve the use of a fixed training budget. We test this by comparing standard Transformers, larger matched-FLOPs Transformers, Looped Transformers \cite{dehghani2019universal}, RINS \cite{alabdulmohsin2026recursive}, and HRM under matched training compute.

\begin{table}[htbp]
\centering
\footnotesize

\newcommand{\thd}[1]{\scriptsize\textbf{#1}}

\begin{tabularx}{\textwidth}{@{} ll p{1.6cm} *{9}{X} @{}}
\toprule
\thd{Model} & \thd{Recursions} & \thd{FLOPs ($10^{21}$)} & \thd{Tokens (T)} & \thd{MMLU} & \thd{\mbox{ARC-C}} & \thd{Hella.} & \thd{Wino.} & \thd{BoolQ} & \thd{DROP} & \thd{GSM8K} & \thd{MATH} \\ 
\midrule
HRM 1B & 4 (H2L3) & 1.0 & 0.06 & \textbf{60.73} & \textbf{81.91} & \textbf{63.42} & \textbf{72.38} & \textbf{86.18} & \textbf{82.21} & \textbf{84.53} & \textbf{56.16} \\ 
\midrule
Looped Transformer 1B & 4 & 1.0 & 0.06 & 56.51 & 74.06 & 47.59 & 68.19 & 82.35 & 76.20 & 75.13 & 48.30 \\
RINS 1B & 4 ($r=7$) & 1.1 & 0.06 & 56.09 & 76.71 & \underline{52.49} & \underline{68.35} & 83.98 & \underline{79.92} & \underline{77.71} & 48.90 \\
\midrule
Transformer 1B & 1 & 1.0 & 0.17 & 53.15 & 74.32 & 47.29 & 65.51 & 83.58 & 75.30 & 75.06 & 48.36 \\
Transformer 3B (Deep) & 1 & 1.1 & 0.06 & \underline{56.67} & \underline{80.46} & 51.05 & 67.96 & \underline{84.07} & 76.95 & 75.66 & \underline{50.50} \\
Transformer 3B (Wide) & 1 & 1.1 & 0.06 & 54.53 & 76.54 & 43.77 & 67.09 & 83.55 & 74.04 & 73.01 & 49.74 \\
\bottomrule
\end{tabularx}
\caption{Training FLOPs-matched comparison of recurrent architectures and Transformer models. Bold denotes the highest score in each column, and underline denotes the second highest.}
\label{tab:recurrent_compare}
\end{table}

\Cref{tab:recurrent_compare} compares training-FLOPs-matched recurrent architectures (including HRM, looped Transformers, and RINS) with standard Transformers. For recursive models, the value in the recursions column indicates total compute per forward pass, expressed as a multiple of the compute required if recurrence is not present. For example, H2L3 denotes 2 outer H cycles, with 3 L steps inside each outer cycle, giving $2 \times (3 + 1) = 8$ total H/L module steps. Since each H or L module contains half of the non-embedding parameters of the full HRM recurrent core, this corresponds to $8 \times 0.5 = 4$ recursions in the table. For standard Transformer models, the value is 1.

Looped Transformers and RINS generally outperform Transformer models of the same size, showing that recurrent or looped computation is an effective architectural direction. When compared with a larger Transformer under a matched training-FLOPs budget, however, their advantage is less consistent. HRM is a strong instance of this architecture-design space and performs well against the listed baselines, including the larger deep Transformer.

Within recurrent designs, we further compare HRM with TRM to separate hierarchical dual-timescale recurrence from a shared-parameter dual-timescale recurrent variant.

\begin{table}[htbp]
\centering
\footnotesize

\newcommand{\thd}[1]{\scriptsize\textbf{#1}}

\begin{tabularx}{\textwidth}{@{} ll p{1.6cm} *{9}{X} @{}}
\toprule
\thd{Model} & \thd{Recursions} & \thd{FLOPs ($10^{21}$)} & \thd{Tokens (T)} & \thd{MMLU} & \thd{\mbox{ARC-C}} & \thd{Hella.} & \thd{Wino.} & \thd{BoolQ} & \thd{DROP} & \thd{GSM8K} & \thd{MATH} \\ 
\midrule
HRM 1B & 4 (H2L3) & 1.0 & 0.06 & \textbf{60.73} & \textbf{81.91} & \textbf{63.42} & \textbf{72.38} & \textbf{86.18} & \textbf{82.21} & \textbf{84.53} & \textbf{56.16} \\ 
TRM 1B (\textit{unstable}) & 2 (H2L1) & 1.1 & 0.06 & 46.38 & 56.66 & 35.22 & 59.35 & 74.74 & 63.56 & 67.63 & 44.12 \\
\midrule
HRM 0.6B & 4 (H2L3) & \textbf{0.5} & 0.06 & 56.87 & 74.32 & 50.48 & 66.38 & \underline{84.86} & 78.07 & 78.39 & 50.28 \\
TRM 0.6B & 4 (H2L3) & 1.1 & 0.06 & \underline{56.88} & \underline{76.71} & \underline{55.12} & \underline{67.01} & 84.22 & \underline{79.91} & \underline{78.47} & \underline{50.84} \\
\bottomrule
\end{tabularx}
\caption{\textbf{Performance and stability comparison against TRM.} HRM maintains stable training dynamics across all scales, whereas TRM suffers from severe instability at the 1B parameter scale. Furthermore, at the 0.6B scale, HRM achieves competitive performance across most benchmarks while requiring 2$\times$ less compute than TRM.}
\label{tab:trm_compare}
\end{table}

TRM is a HRM-variant that shares the H and L module parameters, to achieve strong results on symbolic reasoning problems at smaller scale~\citep{TRM2025}. \Cref{tab:trm_compare} compares HRM and TRM. Since TRM shares parameters across H-L modules, there are two ways to approximately match FLOPs: keeping the overall parameter count fixed and reduce the number of recursions, or keeping the recursive structure fixed and reduce the parameter count. In the first setting, TRM training is less stable, likely due to the reduced recursion weakening the intended iterative computation. In the second setting, the additional recursion stabilizes training and improves performance, but the model still lags behind FLOPs-matched HRM. HRM achieves generally comparable or stronger performance while using substantially fewer FLOPs than TRM in this comparison.

These results support the first part of the small-budget design exploration: recurrent and looped architectures can improve benchmark yield under fixed training compute, and HRM is one effective point in this broader architecture-design space.

\subsection{Task-completion objective and PrefixLM yield}

The second part of this exploration asks whether the training objective and input structure can increase the yield of each training example. We test this through an incremental ablation that starts with a standard Transformer trained on full question--answer pairs using causal attention, then adds the task-completion objective, PrefixLM attention, and finally the HRM architecture. All experiments are FLOPs-matched.

As shown in \Cref{tab:model_comparison}, the task-completion objective, PrefixLM training, and the HRM architecture each significantly contribute to overall performance. Introducing the task-completion objective establishes initial gains across all benchmarks, while PrefixLM training further enhances these results compared to standard causal masking. Ultimately, transitioning from a standard Transformer to the HRM architecture delivers a final, consistent performance increase across the board.

\begin{table}[htbp]
\centering

\newcommand{\thd}[1]{\footnotesize\textbf{#1}}

\begin{tabularx}{\textwidth}{@{} ll p{1.6cm} *{8}{X} @{}}
\toprule
\thd{Architecture} & \thd{Objective} & \thd{Attention} & \thd{MMLU} & \thd{\mbox{ARC-C}} & \thd{Hella.} & \thd{Wino.} & \thd{BoolQ} & \thd{DROP} & \thd{GSM8K} & \thd{MATH} \\ 
\midrule
\multirow{3}{*}{Transformer 1B} & $P(x)$    & Causal   & 40.55 & 51.91 & 32.50 & 48.54 & 39.94 & 38.24 & 48.37 & 35.44 \\
                             & $P(x_a|x_q)$  & Causal   & 47.72 & 62.88 & 41.64 & 60.85 & 76.57 & 54.24 & 69.75 & 47.04 \\
                             & $P(x_a|x_q)$  & PrefixLM & 53.15 & 74.32 & 47.29 & 65.51 & 83.58 & 75.30 & 75.06 & 48.36 \\ 
\midrule 
\multirow{3}{*}{HRM 1B}                         & $P(x)$  & Causal   & 43.68 & 60.24 & 45.10 & 53.71 & 55.98 & 42.74 & 66.19 & 44.32 \\
                             & $P(x_a|x_q)$  & Causal & 50.60 & 69.80 & 50.43 & 63.93 & 67.28 & 62.39 & 79.91 & 54.18 \\ 
                          & $P(x_a|x_q)$  & PrefixLM & \textbf{60.73} & \textbf{81.91} & \textbf{63.42} & \textbf{72.38} & \textbf{86.18} & \textbf{82.21} & \textbf{84.53} & \textbf{56.16} \\ 
\bottomrule
\end{tabularx}
\caption{Performance Comparison across Model Architectures and Objectives}
\label{tab:model_comparison}
\end{table}

\subsection{Comparison with contemporary open models}

After exploring architecture, objective, and input structure under the small-budget setting, we compare the resulting HRM-Text checkpoint with contemporary fully open and open-weight models trained with substantially larger budgets.

\Cref{fig:benchmark_bars} and \Cref{tab:eval_open_weight} compares HRM-Text 1B with contemporary fully open and open-weight models, including Llama, Qwen, Gemma, OLMo and recurrent models, Huginn and Ouro. HRM-Text achieves strong performance among these models on most benchmarks, while remaining competitive on MMLU despite its smaller parameter count and limited 40B unique-token pretraining budget. This pattern is consistent with the role of HRM-Text: recurrent depth and task-completion pretraining improve reasoning and task execution, while broad factual-knowledge coverage remains more sensitive to model scale and data breadth.
HRM-Text reaches this performance range with $96\text{-}432\times$ less estimated training compute and roughly $100\text{-}900\times$ fewer training tokens than the compared open baselines. This comparison supports the paper's central question by showing that a small, task-completion-oriented pretraining run can enter the performance range of open models trained with far larger token and compute budgets.

\begin{table}[htbp]
\centering
\footnotesize

\newcommand{\thd}[1]{\scriptsize\textbf{#1}}

\begin{tabularx}{\textwidth}{@{} ll p{1.6cm} *{9}{X} @{}}
\toprule
\thd{Model} & \thd{Architecture} & \thd{FLOPs($10^{21}$)} & \thd{Tokens(T)} & \thd{MMLU} & \thd{\mbox{ARC-C}} & \thd{Hella.} & \thd{Wino.} & \thd{BoolQ} & \thd{DROP} & \thd{GSM8K} & \thd{MATH} \\ 
\midrule
\multicolumn{11}{@{}l}{\textit{Fully open}} \\
\midrule
HRM-Text 1B & Recurrent & 1 & 0.06 & 60.7 & \textbf{81.9} & 63.4 & \textbf{72.4} & \textbf{86.2} & \textbf{82.2} & \textbf{84.5} & \textbf{56.2} \\ 
Huginn 3.5B & Recurrent & 127 & 0.8 & 31.4 & 38.2 & 65.2 & 59.4 & 69.8 & 17.8 & 34.6 & 12.6 \\
Olmo3 7B & Dense & 252 & 6 & \underline{65.8} & \underline{81.6} & 72.7 & 64.6 & \underline{85.4} & \underline{71.5} & 75.5 & 40.0 \\
\midrule
\multicolumn{11}{@{}l}{\textit{Open weight}} \\
\midrule
Llama3.2 3B & Dense & 162 & 9 & 58.0 & 69.1 & 47.1 & 52.4 & 76.2 & 45.2 & \underline{77.7} & \underline{48.0} \\
Gemma3 4B & Dense & 96 & 4 & 59.6 & 56.2 & \textbf{77.2} & 64.7 & 72.3 & 60.1 & 38.4 & 24.2 \\
Qwen3.5 2B & Dense & 432 & 36 & 64.5 & 81.0 & 64.6 & 56.7 & 80.5 & 30.8 & 53.0 & 34.2 \\
Ouro 1.4B & Recurrent & 259 & 7 & \textbf{67.4} & 60.9 & \underline{74.3} & \underline{72.3} & 83.6 & 49.7 & \underline{78.9} & 22.4 \\ 
\bottomrule
\end{tabularx}
\caption{Evaluation results of HRM-Text 1B and contemporary fully open or open-weight models.}
\label{tab:eval_open_weight}
\end{table}

Our reported scaling experiments extend to 3B parameters for Transformers and 1B parameters for HRM-Text. Within this range, the results show that models trained with a limited amount of data can remain competitive with contemporary industrial-scale pretraining efforts that use much larger datasets (up to 36T tokens). Demonstrating similar efficiency gains at larger model scales remains in the scope of future work.

\subsection{Effective depth analysis}

We hypothesize that HRM's effectiveness is due to its recurrence, increasing the amount of useful internal computation. We test this hypothesis by examining whether HRM exhibits greater effective depth than standard and looped Transformer baselines.

\begin{figure}[h]
  \centering
  \includegraphics[width=\linewidth]{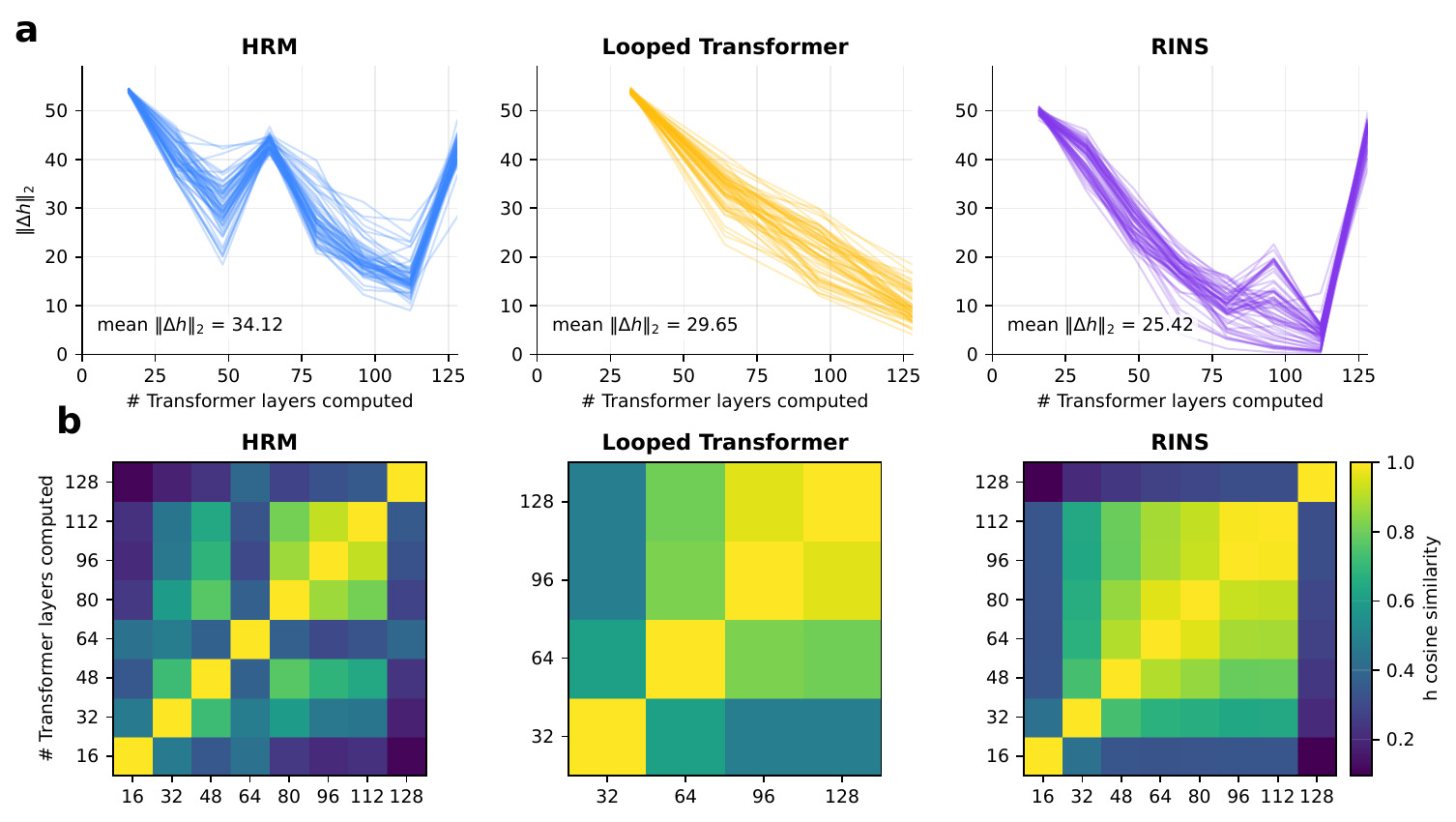}
  \caption{\textbf{Effective depth analysis}. (a) Each layer of HRM consistently reveals considerable changes compared to its previous layer, showing that deep layers of HRM are still making meaningful contributions to the hidden states. (b) HRM has smaller cosine similarity of block-wise representations, while other model variants suffer more from the common layer representation over-smoothing issue, analogously to standard transformers.}
  \label{fig:recurrent_layer_eff}
\end{figure}

\begin{figure}[h]
    \centering
    \includegraphics[width=0.9\textwidth]{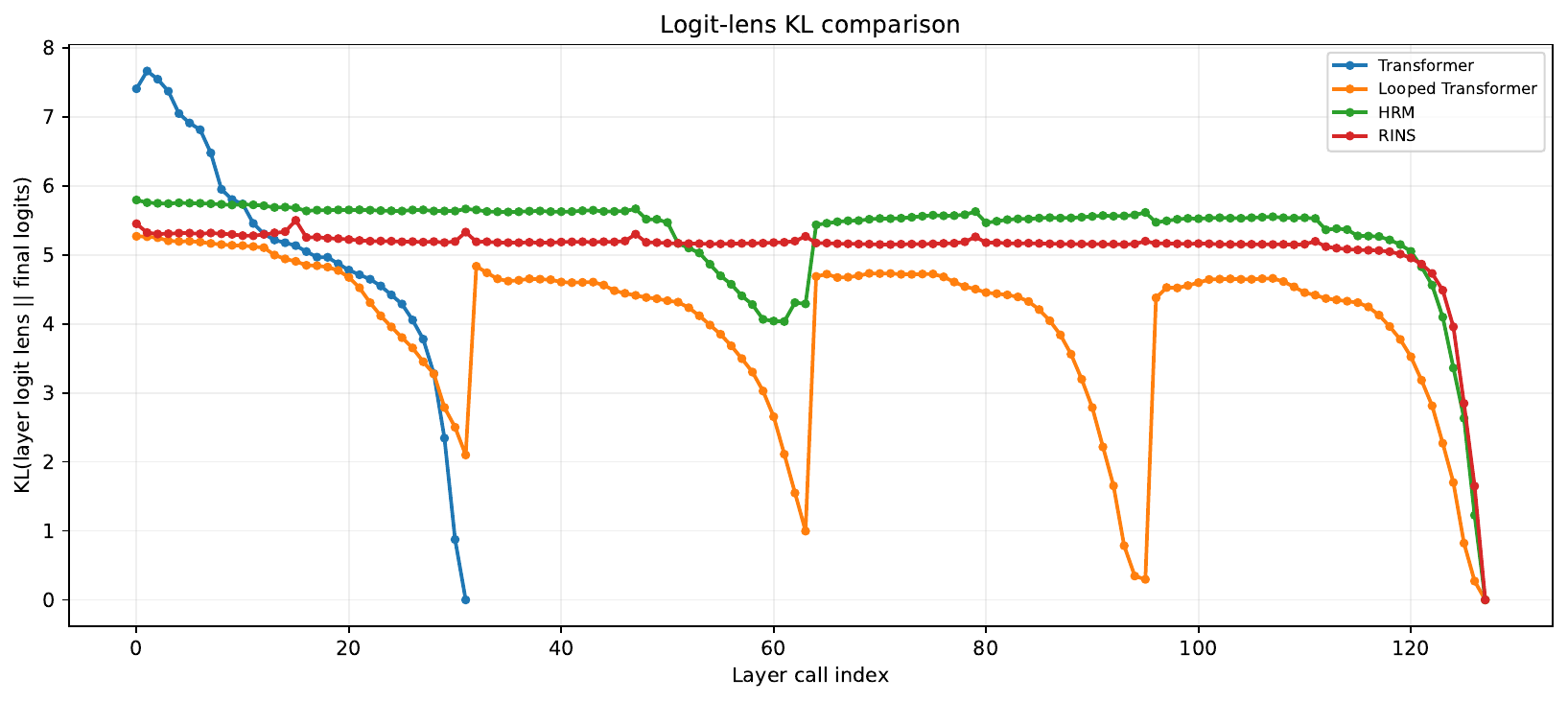}
    \caption{\textbf{Per-layer logit lens KL}. HRM shows the largest logit len KL in deep layers, while both standard and looped transformers converges to stable distributions in shallow layers.}
    \label{fig:logit_lens_kl}    
\end{figure}

\Cref{fig:recurrent_layer_eff} illustrates effective depth from two perspectives: (a) the norm of the difference between adjacent recurrent blocks, and (b) the cosine similarity of block-wise representations.
Both metrics suggest that HRM maintains more active representational change across depth than standard Transformers and other looped models.

Following \citet{hu2025affects}, we also use logit lens analysis to evaluate how early the model's output distribution begins to stabilize. We decode hidden states from different layers using the model's output projection head, then compute the KL divergence between each probed prediction and the final model distribution. As shown in \Cref{fig:logit_lens_kl}, both the standard Transformer and looped Transformer converge to a stable output distribution in relatively early layers, suggesting that their deeper layers make smaller incremental contributions. In contrast, HRM retains larger KL values in deeper layers, indicating greater effective depth.

\section{Training details}
\label{sec:training_details}

\subsection{Dataset}
\label{sec:methods:data}

We train HRM-Text exclusively on open-source datasets, comprising general instructions, rewritten knowledge, mathematical and symbolic tasks, textbook exercises, and web-extracted questions. The initial corpus contains approximately \(176.5\)B tokens across \(593.7\)M documents. From this, we sample \(40\)B unique tokens for a total training duration of \(60\)B tokens, with repetition governed by the stratified sampling schedule described below. Table~\ref{tab:data} summarizes the dataset composition.

\newcolumntype{Y}{>{\raggedright\arraybackslash}X}
\newcolumntype{L}[1]{>{\raggedright\arraybackslash}p{#1}}

\begin{table}[t]
\footnotesize
\centering
\renewcommand{\arraystretch}{1.1}
\setlength{\tabcolsep}{4pt}
\begin{tabularx}{\linewidth}{@{}L{0.22\linewidth} Y r r L{0.15\linewidth}@{}}
\toprule
Type & Datasets & Tokens & Docs & Condition \\
\midrule
General instructions & FLAN \citep{longpre2023flan}, Tasksource \citep{sileo2024tasksource}, NoRobots \citep{huggingfaceh4_2023_norobots} & 138.7B & 379.9M & direct / cot \\
Rewritten Wikipedia knowledge & SYNTH \citep{SYNTH} & 21.7B & 60.8M & synth, direct / cot \\
Math and reasoning & Platypus \citep{lee2023platypus}, Principia \citep{aggarwal2026principia}, OpenMathInstruct2 \citep{toshniwal2025openmathinstruct2}, NuminaMath \citep{numinamath2024}, OmniMATH \citep{gao2024omnimath} & 6.8B & 25.7M & synth / synth, cot \\
Symbolic & DMMath \citep{saxton2019analysing}, AMPS \citep{hendrycks2021measuring}, Sudoku-Extreme \citep{HRM2025} & 6.2B & 120.7M & direct \\
Reasoning (thinking tokens removed) & AceReason \citep{chen2025acereason}, OpenThoughts2 \citep{guha2025openthoughts} & 2.4B & 3.6M & synth, cot \\
Textbook exercises & TextbookReasoning \citep{fan2025megascience} & 358.1M & 1.2M & synth, cot \\
Extracted web instructions & NaturalReasoning \citep{yuan2025naturalreasoning}, WebInstruct-verified \citep{ma2025generalreasoner}, AMPS-khan \citep{hendrycks2021measuring} & 375.1M & 1.8M & noisy / direct / cot \\
\bottomrule
\end{tabularx}
\caption{Source datasets used for HRM-Text training, grouped by type.}
\label{tab:data}
\end{table}
\begin{wraptable}[13]{r}{0.43\linewidth}
\centering
\footnotesize
\caption{Stratified sampling limits used to construct the training mixture.}
\label{tab:data_sampling}
\begin{tabular}{@{}ll@{}}
\toprule
Source dataset & Sampling cap \\
\midrule
FLAN & 5k docs/task \\
Tasksource & 10k docs/task \\
SYNTH & 10M docs \\
AceReason & 2M docs \\
OpenThoughts2 & 500k docs \\
DM Math & 100k docs/task \\
Sudoku & 1M docs \\
Small datasets ($\leq 50$k docs) & $10\times$ original size \\
Other datasets & Original size \\
\bottomrule
\end{tabular}
\vspace{-3.6em}
\end{wraptable}

To control response properties during inference, we prepend specific condition tags to the instructions based on the target response style. We utilize four primary conditions: \texttt{direct} (answer-only), \texttt{cot} (chain-of-thought), \texttt{synth} (synthetic answer style), and \texttt{noisy} (web-crawl text with uneven formatting). As outlined in the ``Condition'' column of Table~\ref{tab:data}, this approach leverages conditioned training \citep{wang2024openchat,dong2023steerlm} to enable explicit selection of the model's output format at generation time.

To concentrate the training signal on final task completions, we strip all text enclosed within \texttt{<think>}...\texttt{</think>} boundaries prior to training. This eliminates explicit long-CoT traces mostly produced by reinforcement learning with verifiable rewards (RLVR) training~\cite{DeepSeekR1}, aligning with our objective for HRM-Text to rely on its internal hierarchical computation rather than explicit reasoning steps.

We employ SeqIO-style stratified sampling \citep{roberts2023scalingt5xseqio}, treating each dataset or task as an independent stratum rather than sampling uniformly from a pooled corpus. To ensure a balanced training mixture and prevent over-representation of massive datasets, we apply caps on number of documents per task or per dataset, while upsampling smaller datasets. The specific sampling limits and multipliers are detailed in \autoref{tab:data_sampling}.

\subsection{Dataset Contamination}

While our pretraining data originates from widely used public sources, and many enforce decontamination measures, residual contamination may still persist considering the scale of pretraining. To rigorously assess whether our models' benchmark performance artificially benefits from exposure to test examples, we conduct a statistical test adapted from the Llama family~\citep{touvron2023llama}.

We tokenize questions from all evaluated benchmarks (excluding few-shot exemplars) and identify $n$-gram matches against the fully tokenized pretraining corpus. A sample's contamination percentage is defined as the fraction of its tokens participating in these matched $n$-grams.

To determine if contamination inflates performance, we partition the evaluation data into four overlapping subsets based on contamination percentage: ``Clean'' ($< 20\%$), ``Not Clean'' ($\ge 20\%$), ``Not Dirty'' ($< 80\%$), and ``Dirty'' ($\ge 80\%$). For contamination to be deemed significantly impactful, ``clean'' samples must perform demonstrably worse than average, while ``dirty'' samples perform demonstrably better. For each subset of size $k$, we compute the empirical mean performance $\bar{X}$ and the test statistic $Z_k = (\bar{X} - \mu_k) / \sigma_k$, where $\mu_k$ and $\sigma_k$ are the mean and standard deviation of the sampling distribution for size $k$. We conclude that dataset contamination provides a statistically significant performance boost only if $|Z_k| > 2$ across all four subsets.

We applied this test to HRM-Text 0.6B and 1B using $n = 13$ and $n = 20$, on all benchmarks shown in \Cref{tab:eval_open_weight}. HRM-Text 0.6B exhibited no significant contamination in either setting. HRM-Text 1B showed statistical significance on the DROP benchmark for $n = 13$ (as shown in \Cref{tab:drop_contamination}), but not for $n = 20$. Nonetheless, HRM-Text 1B still achieves a score of $81.1$ on the strictly clean subset ($0\%$ average contamination, $5904$ samples) of DROP, indicating strong baseline generalization despite a marginal potential benefit from contamination.

Overall, these analyses show that HRM-Text's benchmark performance is unlikely artificially driven by prior exposure to test examples.

\begin{table}[ht]
\centering
\begin{tabular}{llrrrrr}
\toprule
Dataset & Subset Type & Avg. Contam. \% & $n$ & $\bar{X}$ & $\mu_n$ & $Z_n$ \\
\midrule
\multirow{4}{*}{DROP ($n = 13$)} 
 & Clean     & 0.0  & 5904 & 81.1 & 82.3 & $-2.53$ \\
 & Not Clean & 90.5 & 3632 & 84.2 & 82.3 & $3.23$ \\
 & Not Dirty & 6.1  & 6576 & 80.8 & 82.3 & $-3.25$ \\
 & Dirty     & 97.5 & 2960 & 85.5 & 82.3 & $4.85$ \\
\bottomrule
\end{tabular}
\caption{Contamination analysis results for the DROP dataset on HRM-Text 1B.}
\label{tab:drop_contamination}
\end{table}

\subsection{Architecture and optimization details}

HRM-Text 1B took 46 hours to pretrain on two 8×H100 nodes, costing around \$1,472 (assuming \$2 per H100 hour). We summarize the model, optimization, and infrastructure settings below.

\paragraph{Tokenizer.} We employ a Byte-Pair Encoding (BPE) tokenizer with a vocabulary size of 65,536, trained using the \textit{tokenizers} library.

\paragraph{Model configuration.} Each module is a transformer comprising 16 layers, with a hidden size of 1536 and a head size of 128. We use a context size of 4,096 and RoPE positional encoding with $\theta = 10{,}000$. The RMSNorm $\epsilon$ parameter is set to $10^{-6}$. All models are trained in \textit{bfloat16} precision, and all model weights are initialized using LeCun normal.

\paragraph{Optimization.} We use the Adam-atan2 optimizer~\cite{EXWANLGSKLP-2024} with $\beta_1 = 0.9$, $\beta_2 = 0.95$, and a weight decay of 0.1. The learning rate is linearly warmed up over 2,000 steps and then held constant at $2.2 \times 10^{-4}$. No gradient clipping is applied. The batch size is 196,608 tokens. Rather than applying standard learning rate decay, we maintain an Exponential Moving Average (EMA) of the model weights with a decay rate of 0.9999. Both our final evaluations and the publicly released model weights use this EMA checkpoint.

\paragraph{Infrastructure.} The parallelization framework is based on PyTorch FSDP2~\citep{liang2025torchtitan}. All models are trained in a single continuous run. We do not use intermediate checkpointing, crash recovery, or skip loss spikes.

\section{Discussion}
\label{sec:discussion}

\subsection{Toward decoupling knowledge and reasoning} \label{sec:methods:knowledge_vs_reasoning}

Our results suggest a direction for partially decoupling factual coverage from reasoning computation. HRM-Text is trained on only 40B unique tokens, and explicitly knowledge-oriented sources constitute only a fraction of the task-formatted mixture. Nevertheless, the model achieves strong performance on reasoning-heavy benchmarks such as MATH and GSM8K, while retaining nontrivial performance on broader knowledge benchmarks such as MMLU. This pattern suggests that a compact recurrent model can learn useful task-execution and reasoning behavior without requiring the same degree of broad factual memorization typically associated with trillion-token pretraining.

This observation motivates future systems that separate a compact reasoning core from factual storage. In such systems, recurrent models like HRM-Text could specialize in computation, planning, and task execution, while factual breadth is supplied by curated corpora, retrieval-augmented stores, or learned memory modules. Recent conditional-memory approaches such as Engram point in a related direction: instead of forcing the Transformer backbone to simulate static pattern lookup through dense computation, they introduce scalable memory lookup as a complementary sparsity axis, freeing neural computation for global context integration and reasoning~\citep{cheng2026engram}. HRM-Text does not yet implement retrieval or conditional memory, but its results suggest that combining small recurrent reasoning models with external or learned knowledge stores is a promising direction for future work.

\subsection{Adaptive computation time (ACT)}

\citet{HRM2025} equipped HRM with an adaptive computation module that allows simpler problems to terminate earlier, reducing computation while maintaining near-optimal performance. We do not use this component in HRM-Text in order to keep the design and training procedure simpler, but it remains a promising direction for improving both performance and computational efficiency. The current recurrent schedule provides additional effective serial depth, but it also increases inference-time computation relative to a single-pass Transformer. ACT would allow easy prompts or tokens to halt after fewer recurrent cycles while reserving the full recurrent budget for harder cases, potentially recovering a substantial portion of the inference overhead. We therefore view ACT as a natural complement to HRM-Text's recurrent-depth design: HRM supplies depth when reasoning is needed, while ACT can make that depth conditional rather than fixed.

\subsection{PrefixLM with inference frameworks}

PrefixLM can run inside standard text-generation inference frameworks such as vLLM without requiring a fundamentally different serving stack. The main requirement is custom attention-mask handling during prefilling, so that instruction tokens can attend bidirectionally while response tokens remain autoregressive.

Using a PrefixLM-style attention pattern in multi-turn chat also requires careful KV-cache logic: user tokens need full attention within each user segment, while assistant tokens must preserve causal generation. This is an engineering constraint rather than a conceptual limitation, but it should be addressed explicitly in production inference systems.

\section{Conclusion}

We introduced HRM-Text as an empirical existence proof that highly efficient pretraining is achievable. Inspired by biological multi-timescale processing, we co-designed a hierarchical recurrent architecture with a targeted task-completion objective. This demonstrates that there is at least one model family capable of reaching competitive performance without relying on the massive compute and internet-scale raw text that dominate current paradigms.

By drastically reducing the compute-to-performance ratio, this work opens significant potentials for future research. Foundational pretraining is no longer locked inside highly resourced institutions; it is now computationally accessible to small labs, academic groups, and even individuals. We hope this democratization empowers the broader community to actively explore, train, and innovate on new architectures from scratch.

\section{Related Work}
\label{sec:related_work}

The literature on recurrent neural networks and language modelling is extensive. In this section, we discuss the most relevant papers.

\subsection{Scaling laws and efficient pretraining}

Language-model development is driven by scaling laws and compute-optimal training, which together prescribe jointly increasing parameters, data, and compute~\citep{kaplan2020scaling, hoffmann2022training}. This underlies the dominant recipe: large decoder-only Transformers trained on massive corpora and refined via mid- and post-training~\citep{brown2020language,grattafiori2024llama,liu2025midtrainingbridges,mo2025midtraining}. While this scaling paradigm has produced strong models, it concentrates pretraining among compute-rich organizations, reinforcing a growing compute divide~\citep{besiroglu2024compute,ahmed2020democratization}. HRM-Text instead explores whether improved architectures, objectives, and data curation can shift the cost–performance frontier, complementing scaling laws by increasing per-token and per-FLOP efficiency.

\subsection{Conditional sequence modeling and PrefixLM}

The distinction between modeling conditional answers, \(P_\theta(x_a \mid x_q)\), and full text streams, \(P_\theta(x)\), predates modern LLMs. Early sequence-to-sequence models and encoder–decoder transformers explicitly model outputs conditioned on inputs~\citep{sutskever2014,cho2014learning,bahdanau2015neural,vaswani2017attention}.  T5~\citep{raffel2020exploring} later unified NLP tasks as text-to-text generation, reinforcing this conditional framing. In the instruction-tuning phase of language modeling, NLP datasets are converted into instruction–response pairs, and a mask is often applied so that loss is only computed on the response tokens~\citep{wei2022finetuned,sanh2022multitask}. Scaling approaches like FLAN show that such task formatting improves generalization~\citep{longpre2023flan,chung2024scaling}. 

Decoder-only models concatenate the prompt and response into a single causal stream and predict all tokens. Although scalable, this is inefficient: the prompt is known at inference time, yet training still assigns loss to reconstruct it.

PrefixLM-style objectives bridge decoder-only models and conditional generation: prefix tokens attend bidirectionally, while outputs remain causal~\citep{liu2018generating,dong2019unified,raffel2020exploring,tay2023ul2}. HRM-Text builds directly on this lineage by making conditional modeling the primary pretraining objective, using response-only loss and PrefixLM masking to combine encoder–decoder behavior with decoder-only simplicity.

\subsection{Latent computation and recurrent language models}

A line of work seeks to improve model capability by increasing internal computation rather than just scaling parameters or output tokens. Universal Transformers introduced recurrent depth to self-attention~\citep{dehghani2019universal}, and later recurrent or block-recurrent Transformer variants reuse parameters across steps or layers~\citep{hutchins2022blockrecurrent,chowdhury2025investigating,saunshi2025reasoning}. These approaches echo classic recurrent-network ideas but inherit the challenge of unstable long-range credit assignment~\citep{bengio1994learning,zucchet2024recurrent}.

Recent latent-reasoning approaches refine hidden states internally before emitting an answer~\citep{hao2025training,koishekenov2025encode}. Recurrent-depth language models such as Huginn and looped language models such as Ouro~\citep{zhu2025ouro} scale this idea to language modeling and test-time computation~\citep{geiping2026scaling,zhu2025ouro}. Meanwhile, CCDD~\citep{zhou2026coevolutionary} establishes the connection between looped transformers and continuous diffusion language model with latent reasoning advantages. These works demonstrate that latent recurrence is a promising alternative to purely token-level reasoning, but many still rely on large token budgets, stage-wise training, or extensive test-time recurrence.

HRM-Text builds on the Hierarchical Reasoning Model, which uses a two-timescale recurrent design for symbolic reasoning~\citep{HRM2025}. Like prior work, it emphasizes richer internal computation, but it differs in that it is trained from scratch under a small token budget and uses a hierarchical dual-timescale architecture. Related work such as TRM explore even smaller recursive models~\citep{TRM2025}, suggesting that hierarchy, temporal separation and recurrence can enable useful serial computation, though applying them to language remains challenging due to larger states and broader data.

\subsection{Stable recurrent optimization}

Stability is a key challenge for recurrent-depth language models. In Transformers, normalization placement trades off forward stability and gradient flow: PostNorm stabilizes activations but is harder to optimize at depth, while PreNorm improves gradients but risks residual growth and reduced expressivity~\citep{xiong2020layer,liu2020understanding}. Recurrence intensifies this issue, as repeated transformations create long products of Jacobian-like operators during backpropagation. Prior work shows exact long-horizon credit assignment is often impractical~\citep{bengio1994learning,tallec2017unbiasing}, and studies of random matrix products and neural gradients suggest deep multiplicative paths lead to heavy-tailed, lognormal-like variability~\citep{hanin2018products,chmiel2020neural,hodgkinson2021multiplicative}.

HRM-Text addresses these stability issues using architecture-specific techniques: \textit{MagicNorm} and warm-up for deep credit assignment. These design choices distinguish HRM-Text from generic looped Transformers and are crucial to making recurrent depth stable at language-model scale.

\section*{Acknowledgements}

We thank Sen Song, Jiacheng You, and Andy L. Siy for their insightful discussions.

\clearpage
\begin{hyphenrules}{nohyphenation}
\setlength{\bibsep}{.5ex plus .8ex}
\bibliographystyle{unsrtnat}
\bibliography{main}
\end{hyphenrules}

\clearpage
\appendix
\section*{Appendix}

\section{FLOPs estimation}

For dense models, we use the standard training-FLOPs estimate $F = 6ND$.

For recurrent models, we account separately for the forward and backward recurrent unrolls. We count $2ND$ for forward computation and $4ND$ for backward computation, then scale these terms by the number of recurrent steps included in each pass.

\section{Evaluation details}

\label{appendix:sec:evaluation_details}

\begin{table}[htbp]
    \centering
    \caption{Shared evaluation configuration.}
    \label{tab:evaluation_config}
    \begin{tabular}{ll}
        \toprule
        Setting & Value \\
        \midrule
        Maximum evaluation context & 3072 tokens \\
        Decoding temperature & 0 \\
        System prompt & None \\
        Prompt contents & Original question only \\
        Baseline scores & Original papers when available; otherwise rerun from open weights \\
        \bottomrule
    \end{tabular}
\end{table}

The evaluation prompt contains the original benchmark question and, when required by the benchmark protocol, the corresponding few-shot examples. Few-shot examples are added only for few-shot evaluations. We do not add an additional system prompt. Unless otherwise specified, decoding is deterministic with temperature zero and a maximum context length of 3072 tokens.

Baseline scores are taken from the original papers when those scores are available under comparable settings. When paper-reported numbers are unavailable, we evaluate the corresponding open-weight model directly. All few-shot evaluations are run with the same configuration used for HRM-Text and use the vLLM inference engine. Chain-of-thought evaluations are run with \texttt{lm\_eval\_harness}.

\section{Stable optimization in recurrent-depth models}
\subsection{Gradient Stability Under Deep BPTT in HRM} \label{appendix:sec:bptt}

Backpropagation through time (BPTT) is the canonical mechanism for training recurrent computation graphs, yet extensive prior work has established that propagating gradients through the full unroll is often unnecessary and can be detrimental to optimization. On the other hand, truncating the backward horizon can improve practical convergence by trading exact long-range credit assignment for gradients that behave more favorably as stochastic estimators. This bias--stability tradeoff has been formalized and leveraged in the recurrent literature, motivating both principled truncation schemes and analyses of when and why truncation improves training dynamics \citep{tallec2017unbiasing}. In contrast, analogous diagnostics remain underdeveloped for HRM, or other modern looped architectures, where repeated application of a shared block induces an implicit recurrence and the effective backward depth is controlled by the number of backpropagated loop iterations.

\begin{figure}[t]
    \centering
    \begin{subfigure}{0.49\linewidth}
        \centering
        \includegraphics[width=\linewidth]{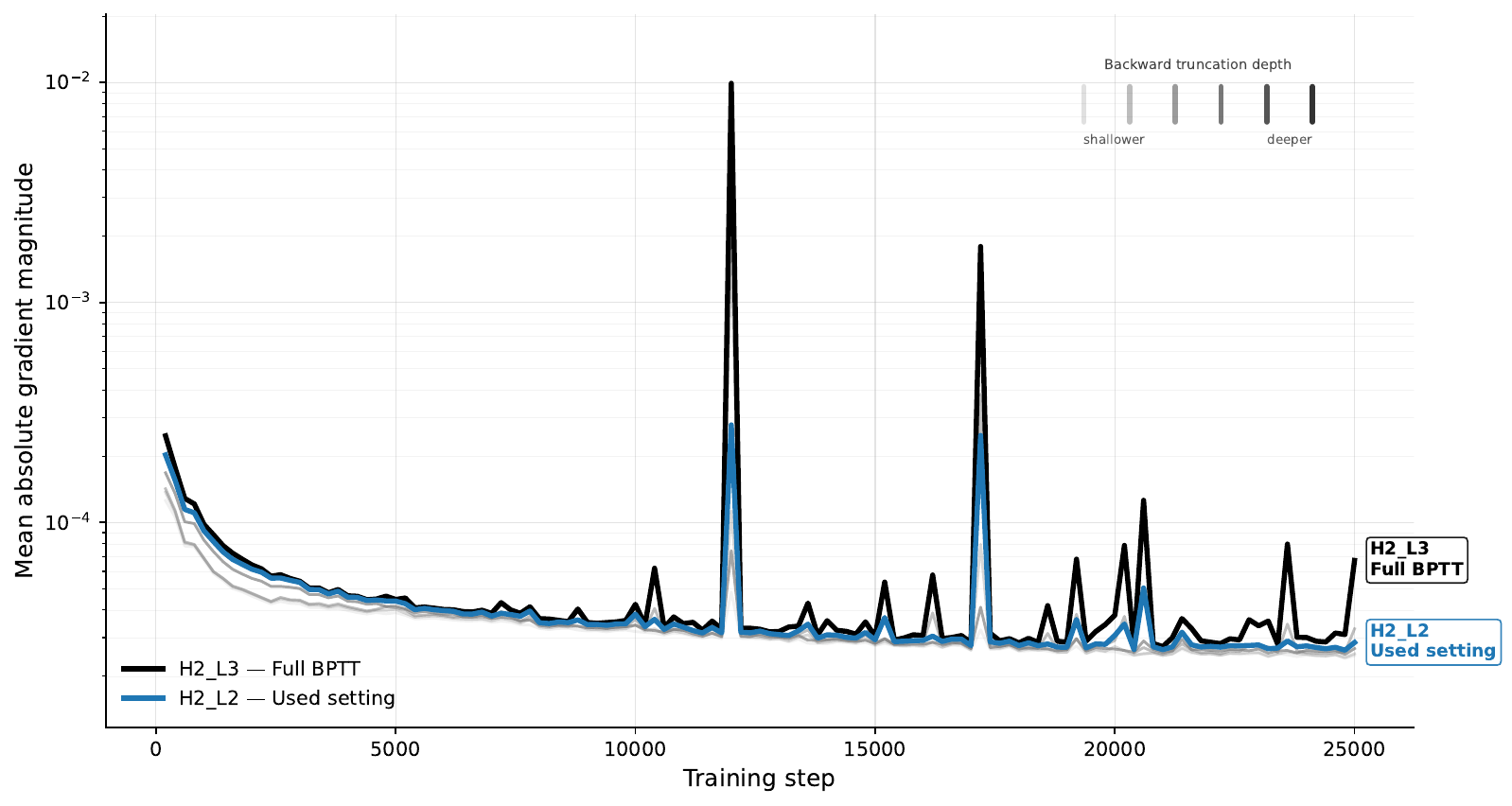}
        \vspace{5pt}
        \caption{Mean absolute gradient magnitude over training.}
    \end{subfigure}
        \hfill
    \begin{subfigure}{0.49\linewidth}
        \centering
        \includegraphics[width=\linewidth]{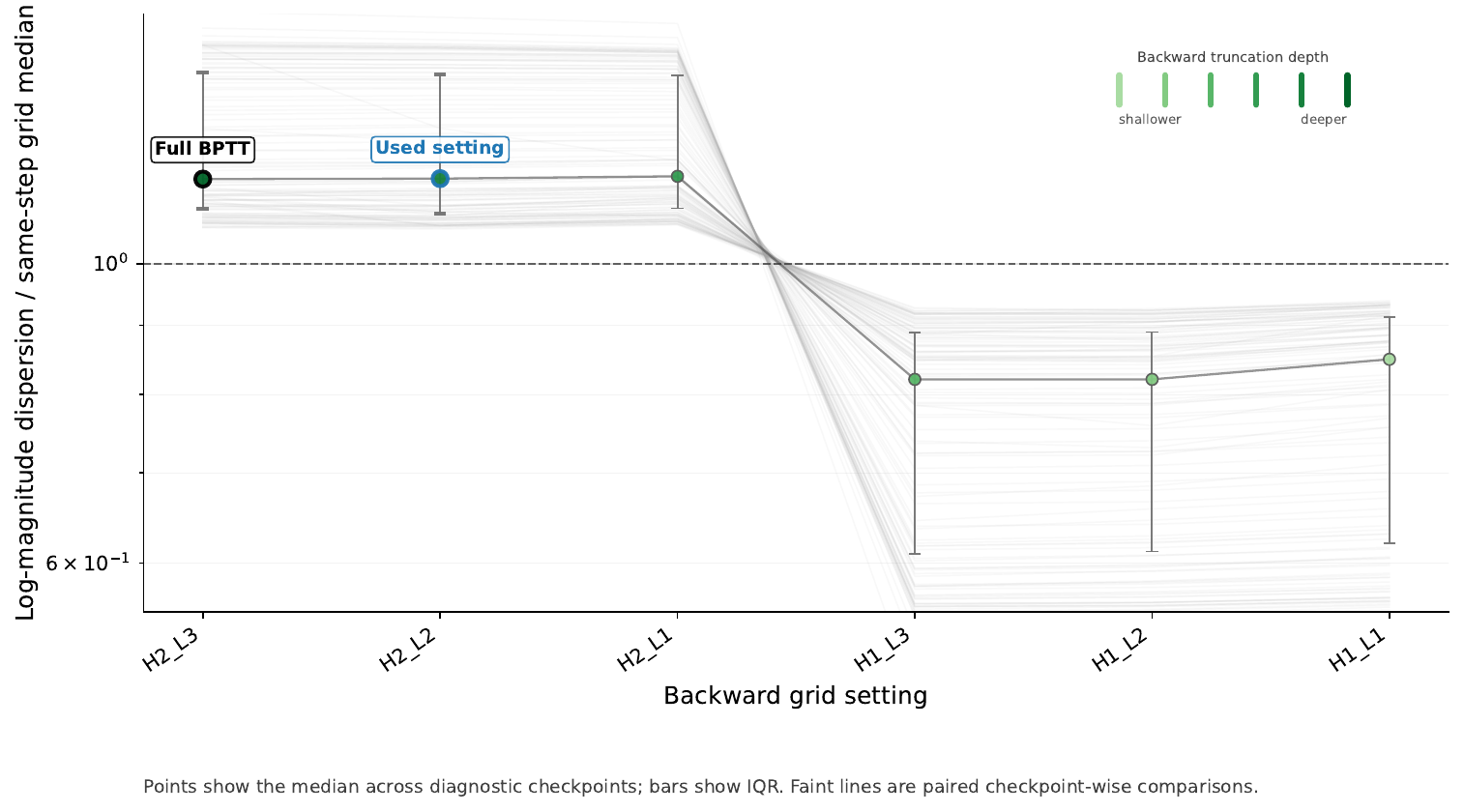}
        \caption{Log-magnitude dispersion.}
    \end{subfigure}
    \caption{
    (a) Full BPTT exhibits rare but substantially larger gradient-magnitude spikes compared with the truncated setting, suggesting that longer backward horizons introduce intermittent high-amplitude gradient events. (b) Values are normalized within diagnostic checkpoints to isolate the effect of backward depth from global training-time drift. Deeper H cycling increases log-magnitude dispersion}
    \label{fig:bptt_dispersion}
\end{figure}

We hypothesize that the instabilities observed under deep BPTT in looped architectures are a consequence of the intrinsically multiplicative structure of gradient propagation through repeated iterations. Specifically, gradients backpropagate through products of Jacobian-like operators across loop steps, and theory for products of many random matrices predicts that the logarithm of the norm of such products is approximately Gaussian, implying lognormal-like variability in gradient magnitudes and increasing separation between typical and extreme values as backward depth grows \citep{hanin2018products}. Complementing this theoretical perspective, empirical evidence suggests that neural gradient magnitudes are often close to lognormal, consistent with multiplicative mechanisms that concentrate mass near small magnitudes while producing comparatively heavier tails \citep{chmiel2020neural,hodgkinson2021multiplicative}.

To test these hypotheses, we perform a targeted study of gradient dynamics in HRM as we systematically increase the number of backward H and L cycles while holding the forward computation fixed. We first quantify instability using the \emph{mean absolute gradient magnitude} and show in Figure~\ref{fig:bptt_dispersion}a that extending the backward horizon toward full BPTT yields substantially more intermittent high-amplitude gradient events over training. We then characterize distributional heterogeneity using a complementary dispersion measure in Figure~\ref{fig:bptt_dispersion}b, which reports \emph{log-magnitude dispersion}, defined as $\mathrm{Std}(\log(|g|+\varepsilon))$. 
This measure supports the view that deeper backward cycling increases multiplicative heterogeneity in gradient magnitudes. Notably, the increase in log-magnitude dispersion is driven primarily by the H-cycle depth rather than the L-cycle depth. 
We therefore interpret the H dimension as the dominant contributor to gradient-magnitude spread in these experiments.

\begin{figure}[t]
    \centering
    \includegraphics[width=\linewidth]{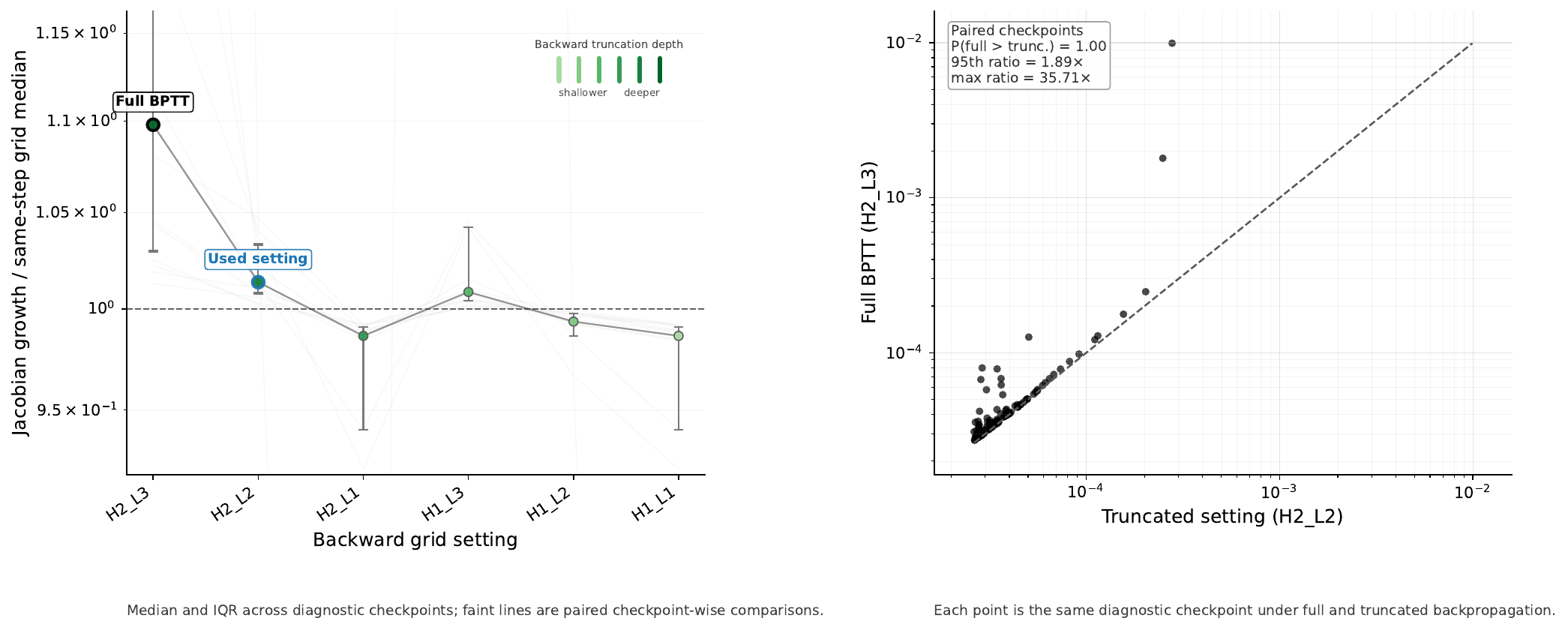}
    \caption{\textbf{Mechanistic evidence for multiplicative gradient instability.}
    Left: Jacobian growth increases with deeper backward cycling, consistent with stronger amplification through products of loop Jacobians. Right: paired full-vs-truncated gradient magnitudes show that full BPTT produces rare, disproportionately large gradient events at the same diagnostic checkpoints.}
    \label{fig:bptt_mechanism}
\end{figure}

Finally, we examine whether the observed gradient spikes are consistent with a multiplicative amplification mechanism. Figure~\ref{fig:bptt_mechanism}a shows that Jacobian growth increases with backward depth, indicating that deeper backpropagation through the recurrent computation amplifies some directions more strongly. Figure~\ref{fig:bptt_mechanism}b provides a paired comparison between the truncated setting used for training and the full-BPTT reference at identical diagnostic checkpoints. The paired scatter shows that full BPTT is often comparable to truncation but occasionally produces much larger gradient magnitudes, consistent with the hypothesis that full backward unrolling primarily harms optimization through rare, high-amplitude tail events rather than through a uniform increase in gradient scale.

The truncation setting used in our experiments is the closest setting to full BPTT that remains stable during training, as illustrated in Figure~\ref{fig:bptt_dispersion}a and Figure~\ref{fig:bptt_mechanism}.

\subsection{Gradient stability across recurrent architectures}

We further compare HRM against RINs and the Universal Transformer through the lens of gradient stability. Since all three architectures reuse computation over depth or recurrence, stable gradient dynamics are an important part of whether the architecture can be trained effectively, rather than merely whether it has sufficient expressive capacity. We therefore evaluate two complementary statistics across runs: the median absolute gradient magnitude and the tail-to-median ratio.

\begin{figure}[t]
    \centering
    \begin{subfigure}[t]{0.48\linewidth}
        \centering
        \includegraphics[width=\linewidth]{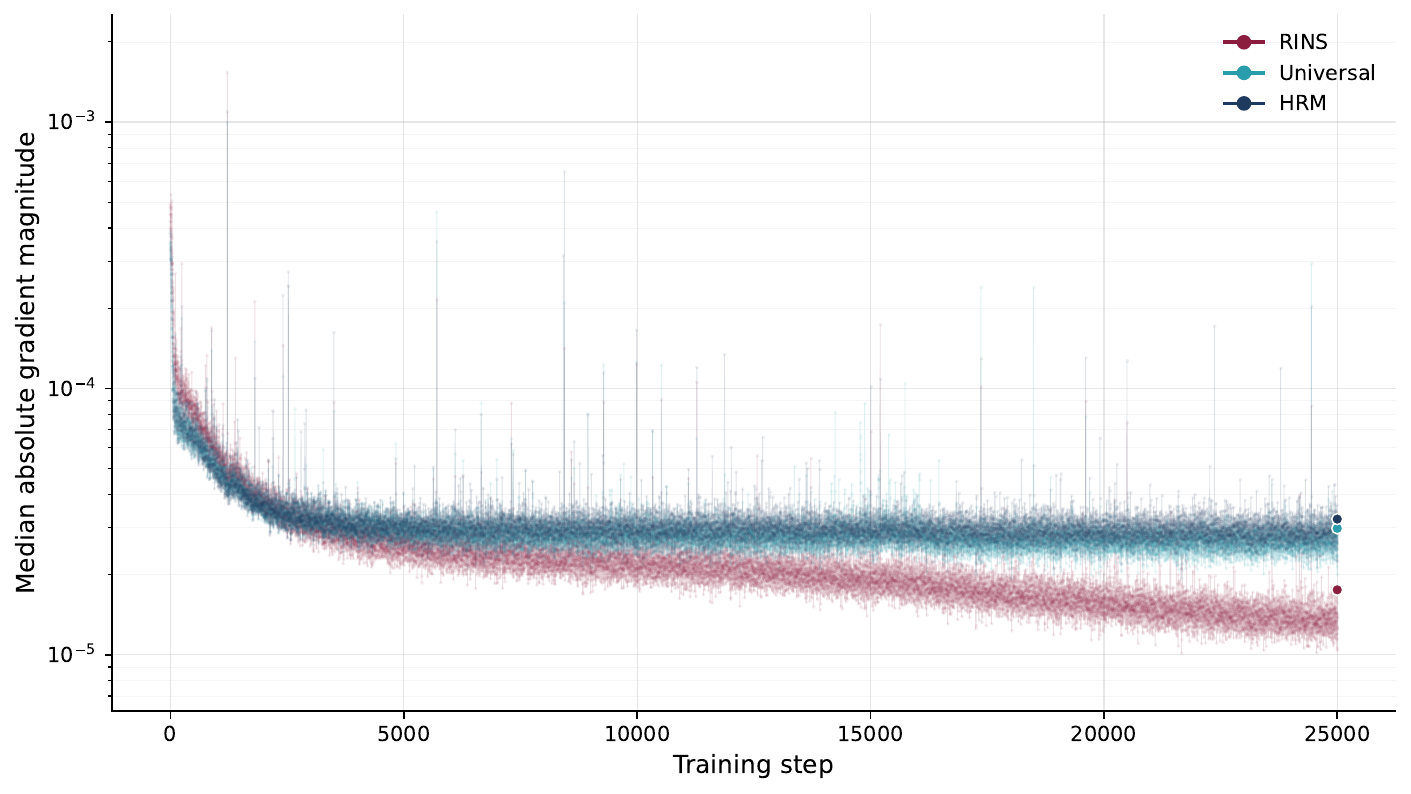}
        \caption{Median absolute gradient magnitude across runs.}
        \label{fig:gradient_abs_p50_across_runs}
    \end{subfigure}
    \hfill
    \begin{subfigure}[t]{0.48\linewidth}
        \centering
        \includegraphics[width=\linewidth]{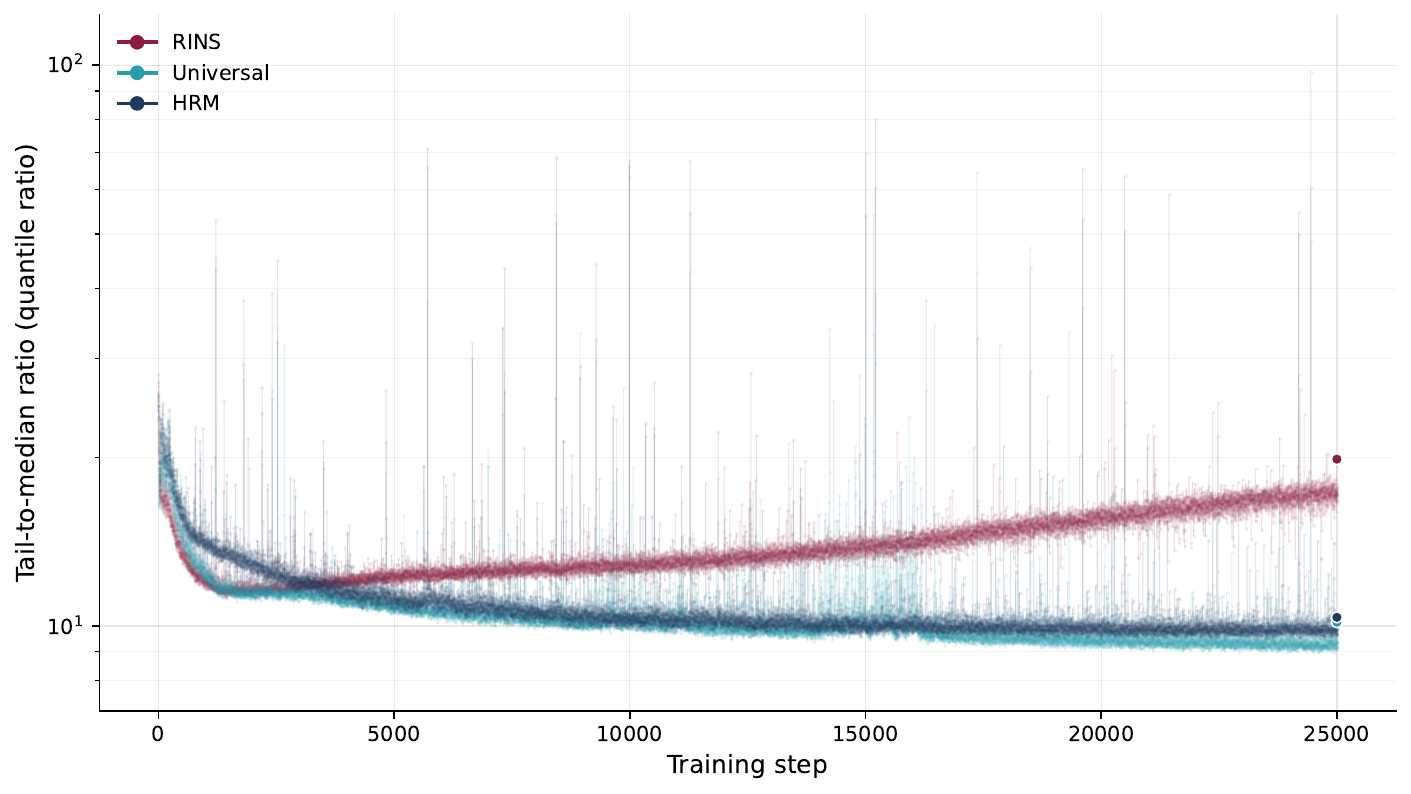}
        \caption{Tail-to-median gradient ratio across runs.}
        \label{fig:gradient_tail_to_median_across_runs}
    \end{subfigure}
    \caption{
    Gradient stability comparison between RINs, HRM, and the Universal
    Transformer. HRM maintains a strong gradient signal while exhibiting
    increasingly even gradient dynamics over training, matching the stability of
    the Universal Transformer more closely than RINs.
    }
    \label{fig:gradient_stability_comparison}
\end{figure}

Figure~\ref{fig:gradient_abs_p50_across_runs} shows that HRM and the Universal Transformer maintain a stronger training signal as optimization progresses: their median absolute gradient magnitudes remain higher than those of RINs across training. At the same time, Figure~\ref{fig:gradient_tail_to_median_across_runs} shows that this signal does not come from increasingly unstable or rare extreme updates. Instead, HRM and the Universal Transformer exhibit lower
tail-to-median ratios over training, indicating that the gradient distribution becomes more even, less heavy-tailed, and less dominated by rare large updates.

This places HRM in the favorable regime of retaining useful gradient signal while avoiding the instability associated with heavy-tailed gradient dynamics. Combined with the main results, this suggests that HRM preserves the training stability of stronger recurrent-depth baselines while delivering better downstream performance.

\section{Inference-time analysis}

\begin{table}[htbp]
\centering
\footnotesize

\newcommand{\thd}[1]{\scriptsize\textbf{#1}}

\begin{tabularx}{\textwidth}{@{} ll p{2.5cm} *{6}{X} @{}}
\toprule
\thd{Model} & \thd{Guidance} & \thd{MMLU} & \thd{\mbox{ARC-C}} & \thd{Hella.} & \thd{Wino.} & \thd{BoolQ}  \\ 
\midrule
HRM 1B & w/o & 61.89 & 80.80 & 62.18 & 73.01 & 88.17 \\
HRM 1B & w/  & \textbf{62.12} & 80.80 & \textbf{62.31} & \textbf{73.17} & \textbf{88.29} \\
& & (w=0.1) & (w=0) & (w=-0.1) & (w=0.5) & (w=-0.5)\\
\bottomrule
\end{tabularx}
\caption{\textbf{Inference-time auto-guidance.} We report the base performance of standard inference, and the best performance along with the corresponding guidance scale $w\in \{-0.5, -0.1, 0, 0.1, 0.5\}$.}
\label{tab:auto_guidance}
\end{table}

At inference time, we enable the auto-guidance~\citep{karras2024guiding} mechanism specifically designed for HRM, which guides itself by interpolating or extrapolating logits from various recursion depths. While having similar motivations, auto-guidance is more efficient than classifier-free guidance (CFG)~\citep{ho2022classifier}: it induces zero computation overhead because the hidden representations from shallow loops are already accessible at decoding time.

In particular, suppose we have the final hidden state $h$ and another hidden state $h'$ from an earlier recurrent loop, both decoded by the LM head. Auto-guidance with guidance scale $w$ is calculated as:
\begin{equation*}
    \text{logits}_w=(1+w)\cdot \text{logits}(h)-w\cdot \text{logits}(h')
\end{equation*}
$w=0$ recovers the standard final prediction; $w>0$ corresponds to extrapolation between the final layer and a shallower layer, treating the shallower prediction as a negative direction; and $w<0$ corresponds to interpolation, where the model balances predictions from shallow and deep recurrent states.

\Cref{tab:auto_guidance} reports HRM performance with and without auto-guidance, where the guidance scale is searched over $w\in\{-0.5, -0.1, 0, 0.1, 0.5\}$. We use an HRM model with two high-level loops and interpolate or extrapolate the logits from these two $H$ modules. Because the intermediate hidden states are already available, auto-guidance introduces no additional computation. It slightly improves performance at test time, and the best guidance scale varies across benchmarks, suggesting that different tasks may benefit from different effective recurrent depths.

Auto-guidance is also closely related to adaptive computation time (ACT) and test-time scaling (TTS). When interpolation ($w<0$) yields better results, the task may not require the full recurrent depth, suggesting that early stopping could improve efficiency. Conversely, when extrapolation ($w>0$) performs better, the task may benefit from deeper recurrent computation and could be a candidate for adaptive test-time scaling. The results in \Cref{tab:auto_guidance} therefore suggest that HRM inference can support adaptive control of recurrent depth, balancing efficiency and performance at test time.

\end{document}